%% file: paper.tex
\documentclass{article}

 \usepackage[preprint]{neurips_2026}

\usepackage[utf8]{inputenc} 
\usepackage[T1]{fontenc}    
\usepackage{hyperref}       
\usepackage{url}            
\usepackage{booktabs}       
\usepackage{amsfonts}       
\usepackage{nicefrac}       
\usepackage{microtype}      
\usepackage{xcolor}         
\usepackage{amsmath} 
\usepackage{amssymb}
\usepackage{wrapfig}

\usepackage{booktabs}
\usepackage{multirow}
\usepackage{array}
\usepackage{xcolor}
\usepackage{colortbl}
\usepackage{graphicx}
\definecolor{ourrow}{HTML}{E8E8E8}
\title{Heuristics are Not Enough: Calibrated and Corrected Caching for Video Diffusion Acceleration}

\author{
\bf Jiyue Jiang$^{\heartsuit}$$^{\ast}$,
Zhiyu Huo$^{\heartsuit}$\thanks{These authors are co-first authors.},
Yuhan Chen$^{\heartsuit}$,
Yanruisheng Shao$^{\heartsuit}$,
Gong Zhang$^{\spadesuit}$,
Lei Deng$^{\spadesuit}$\thanks{Corresponding author}\\
$^{\heartsuit}$ Department of Computer Science and Engineering, The Chinese University of Hong Kong\\
$^{\spadesuit}$ Theory Lab, 2012 Labs, Huawei Technology Co. Ltd\\
{\tt
\{jiangjy, zhiyuhuo, yuhanchen, 1155231342\}@link.cuhk.edu.hk
}\\
{\tt
\{nicholas.zhang, deng.lei2\}@huawei.com
}\\
}

\begin{document}

\maketitle

\begin{abstract}
Feature caching has become a mainstream training-free strategy for accelerating video diffusion models by reusing intermediate outputs across denoising steps. However, existing adaptive methods rely on heuristic difference measures misaligned with actual generation quality, employ a single global threshold that ignores heterogeneous phase dynamics, and never validate their budget estimates against true approximation error, allowing systematic bias to propagate uncorrected. We propose \textbf{AlignCache}, a closed-loop caching framework that directly aligns scheduling decisions with their true influence on the final generated video through three tightly coupled components: (1) phase-adaptive budget routing that partitions the denoising trajectory into structurally distinct phases, each governed by an independent threshold and a staleness-aware guard; (2) closed-loop threshold calibration that exploits counterfactual error measurement at the classifier-free guided output level to adjust decision boundaries online via an EMA feedback loop; and (3) momentum-augmented draft synthesis combining direction-magnitude gated residual extrapolation with anchor-guided unconditional correction. By transforming scheduling from a static risk accumulator into a feedback-driven, self-correcting controller, AlignCache consistently achieves superior acceleration-quality trade-offs across state-of-the-art video generation models. Our code is available at \url{https://github.com/jiangjyjy/AlignCache}.
\end{abstract}

\section{Introduction}

Diffusion models~\cite{ho2020denoising, peebles2023scalable, rombach2022high} have become the dominant paradigm for high-quality generative modeling, with Diffusion Transformers (DiTs)~\cite{peebles2023scalable} powering state-of-the-art video generation systems such as Open-Sora~\cite{zheng2024open}, HunyuanVideo~\cite{kong2024hunyuanvideo}, and Wan2.1~\cite{wang2025wan}. Despite their impressive generation quality, the iterative denoising nature of these models incurs substantial computational costs, dozens of sequential forward passes through billion-parameter networks, making efficient inference a critical bottleneck for deployment and motivating a growing body of training-free acceleration research.

Among various training-free strategies, feature caching~\cite{liu2025timestep, lv2024fastercache, zhao2024pab} has emerged as an effective paradigm by exploiting temporal redundancy across denoising steps. Static caching methods~\cite{selvaraju2024fora, zhao2024pab} adopt fixed intervals but cannot adapt to timestep dynamics, while adaptive approaches~\cite{kahatapitiya2024adaptive, liu2025timestep, lv2024fastercache, wimbauer2024cache, liu2024easycache} use runtime metrics to decide whether to skip or compute. However, existing adaptive methods exhibit three structural limitations: (i)~they rely on heuristic difference measures that do not reflect the impact of caching on the generated video; (ii)~they employ a single global threshold blind to heterogeneous phase dynamics; and (iii)~their budget estimates are never validated against actual approximation error, allowing systematic bias to propagate uncorrected during sampling.

\begin{figure}[t]
    \centering
    \includegraphics[width=\textwidth]{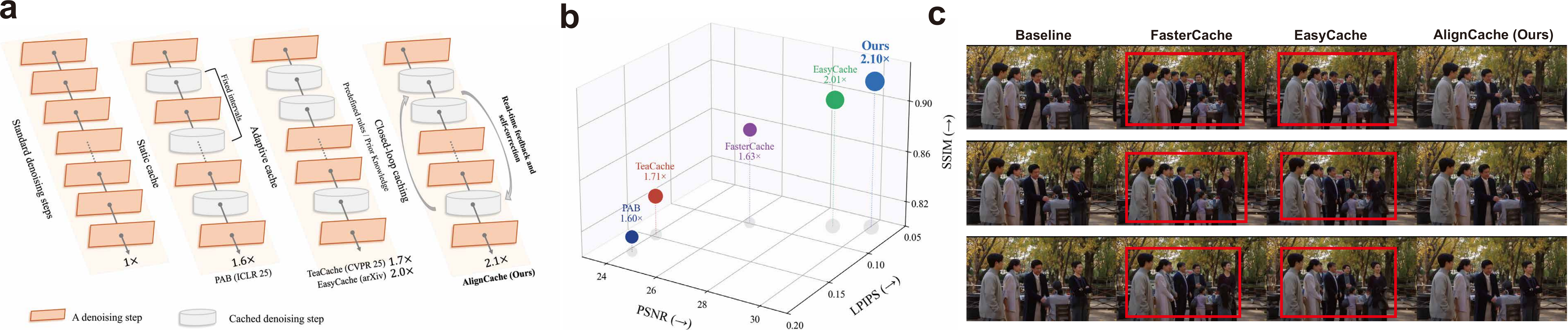}
    \caption{\textbf{Overview and motivation of AlignCache.}
\textbf{(a)} Existing schedulers rely on predefined rules and prior knowledge to allocate exact and cached steps, while AlignCache leverages real-time feedback and self-correction.
\textbf{(b)} AlignCache dominates other methods on PSNR, LPIPS, and speedup.
\textbf{(c)} Comparison on \emph{``The bund Shanghai, pan left.''}: AlignCache preserves fine details where competing methods distort (red boxes). Videos in (c) are generated by HunyuanVideo.}
    \label{fig:small}
\end{figure}

In this work, we propose \textbf{AlignCache}, a closed-loop caching framework that directly aligns scheduling decisions with their actual influence on the final generated video. As illustrated in Figure~\ref{fig:small}, in contrast to prior schedulers that rely on predefined rules and prior knowledge to dispatch exact and cached denoising steps, AlignCache leverages real-time feedback and self-correction to achieve a superior speed-quality trade-off while preserving fine-grained visual details that competing methods tend to distort. AlignCache introduces three tightly coupled components: (1)~a \emph{phase-adaptive budget routing} mechanism that partitions the denoising trajectory into structurally distinct phases (early, mid, late), each governed by an independent threshold and a staleness-aware guard; (2)~a \emph{closed-loop threshold calibration} module that exploits counterfactual observations at each budget-triggered exact computation to measure the true approximation error at the classifier-free guided output level, and dynamically adjusts the decision boundary via an exponential moving average feedback loop; and (3)~a \emph{momentum-augmented draft synthesis} strategy that employs scaled residual extrapolation with anchor-guided unconditional correction, improving skip-mode fidelity and enabling more aggressive caching without sacrificing quality. By transforming the scheduling paradigm from a static, open-loop risk accumulator into a feedback-driven, self-correcting controller, AlignCache achieves superior acceleration-quality trade-offs across diverse video generation models.

Our main contributions are: (1)~We identify the misalignment between heuristic caching criteria and generation quality, and propose AlignCache, a closed-loop framework bridging this gap via counterfactual error measurement. (2)~We design a phase-adaptive budget routing mechanism with staleness-aware guards and closed-loop threshold calibration, replacing static thresholds with a self-correcting controller. (3)~We introduce a momentum-augmented draft synthesis combining residual extrapolation with anchor-guided correction. (4)~Experiments on multiple video generation models demonstrate that AlignCache consistently achieves superior acceleration-quality trade-offs.

\section{Related Works}

\subsection{Video Diffusion Models}

Diffusion models~\cite{ho2020denoising, peebles2023scalable, rombach2022high} have established themselves as the dominant paradigm for generative modeling. Early video generation efforts built upon U-Net architectures~\cite{ronneberger2015u} and latent diffusion frameworks~\cite{blattmann2023stable, blattmann2023align, hong2023cogvideo} to synthesize temporally coherent content. More recently, the community has shifted towards Diffusion Transformers (DiTs)~\cite{peebles2023scalable} for their superior scalability in modeling complex spatiotemporal dynamics. State-of-the-art DiT-based systems including Sora~\cite{brooks2024video}, Open-Sora~\cite{zheng2024open}, CogVideoX~\cite{yang2024cogvideox}, HunyuanVideo~\cite{kong2024hunyuanvideo}, and Wan2.1~\cite{wang2025wan} have achieved remarkable fidelity and coherence, largely benefiting from Rectified Flow~\cite{liu2022flow} formulations that enable high-quality generation with fewer sampling steps. Despite this progress, the iterative denoising nature of these models still incurs substantial computational costs, making efficient inference a pressing challenge for practical deployment.

\subsection{Feature Caching for Diffusion Acceleration}

To accelerate diffusion inference without retraining, various training-free strategies have been explored, including ODE solvers~\cite{karras2022elucidating, lu2022dpm, song2021denoising}, token merging~\cite{bolya2023token, zhang2025training, zou2025accelerating}, and sparsity-based approaches~\cite{wang2024attention, xi2025sparse, ye2024training}. Among these, feature caching has emerged as an effective paradigm for video generation by exploiting temporal redundancy across denoising steps. Static caching methods~\cite{chen2024delta, selvaraju2024fora, zhao2024pab} use fixed intervals and cannot adapt to timestep importance, while adaptive caching approaches~\cite{chu2025omnicache, kahatapitiya2024adaptive, liu2025timestep, lv2024fastercache, wimbauer2024cache} introduce runtime metrics for dynamic decisions yet rely on heuristic difference measures that do not reflect the impact of caching on generation quality, leading to either conservative caching with limited speedup or aggressive caching with noticeable quality degradation. Recent works~\cite{liu2025taylorseer, wang2026precisecache} address this misalignment through frequency decomposition or higher-order feature forecasting, yet the fundamental gap between caching criteria and perceptual quality remains under-explored. In contrast, AlignCache directly aligns the caching decision criterion with the influence on the generated video via closed-loop counterfactual error measurement, enabling more precise redundancy identification and superior acceleration-quality trade-offs.

\section{Preliminaries}
\label{sec:pre}

\begin{figure}[t]
    \centering
    \includegraphics[width=0.97\textwidth]{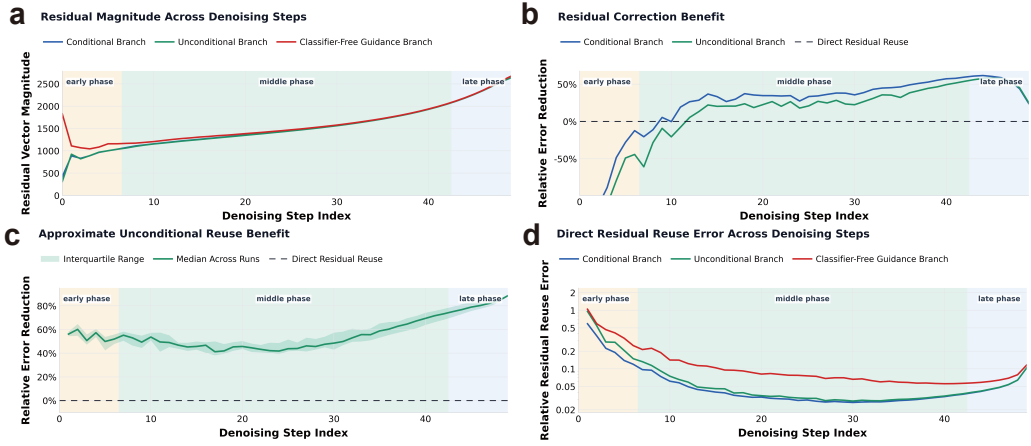}
    \caption{\textbf{Empirical analysis of residual dynamics across denoising steps on Wan2.1-1.3B.}
\textbf{(a)} Residual magnitude exhibits distinct early, middle, and late phases.
\textbf{(b)} Residual correction yields large error reduction over direct reuse in mid-to-late phases.
\textbf{(c)} Approximate unconditional reuse provides consistent benefit across the trajectory.
\textbf{(d)} Direct residual reuse error grows along the trajectory and differs across branches.}
    \label{fig:AlignCache_pre}
\end{figure}

\textbf{Empirical Analysis of Residual Dynamics.}
To motivate AlignCache, we analyze how output residuals evolve along the denoising trajectory of Wan2.1-1.3B (Figure~\ref{fig:AlignCache_pre}). Three observations emerge: (i)~residual magnitude exhibits markedly heterogeneous dynamics across the early, middle, and late phases, making a single global threshold structurally insufficient; (ii)~properly designed correction reduces relative error by $30$-$50\%$ over direct reuse, with the largest gap precisely where direct reuse becomes most error-prone; and (iii)~direct reuse error varies systematically across conditional, unconditional, and CFG-guided branches, indicating that scheduling decisions made on a single branch cannot reliably predict the distortion injected into the final output. These observations directly motivate our phase-adaptive routing~(\S\ref{sec:routing}), momentum-augmented draft synthesis~(\S\ref{sec:draft}), and closed-loop calibration at the CFG-guided level~(\S\ref{sec:calibration}), respectively.

\textbf{Cache-Based Acceleration for Guided Diffusion.}
In classifier-free guidance (CFG)~\cite{ho2022classifier}, each denoising step evaluates the network $\boldsymbol{\epsilon}_\phi$ twice, once on the text prompt $\mathbf{c}$ and once on the unconditional input $\mathbf{c}_{\varnothing}$, forming the guided prediction $\tilde{\boldsymbol{\epsilon}}_t = \boldsymbol{\epsilon}_\phi(\mathbf{x}_t, t, \mathbf{c}_{\varnothing}) + \gamma\big(\boldsymbol{\epsilon}_\phi(\mathbf{x}_t, t, \mathbf{c}) - \boldsymbol{\epsilon}_\phi(\mathbf{x}_t, t, \mathbf{c}_{\varnothing})\big)$ with guidance scale $\gamma$. We index timesteps by $t$, with conditional output $\mathbf{c}_t$ and unconditional output $\mathbf{u}_t$. Cache-based methods~\cite{ma2024deepcache, wimbauer2024cache, liu2024easycache} exploit the temporal smoothness of $\{\mathbf{c}_t, \mathbf{u}_t\}$ by caching the output residual $\mathbf{r}_t^{(b)} = \mathbf{y}_t^{(b)} - \mathbf{x}_t^{(b)}$ for each branch $b \in \{\mathrm{cond}, \mathrm{uncond}\}$ at the most recent \emph{exact anchor} $a(t)$ and reusing it via $\hat{\mathbf{y}}_t^{(b)} = \mathbf{x}_t^{(b)} + \mathbf{r}_{a(t)}^{(b)}$. A \emph{scheduler} decides at every timestep whether to execute a full forward pass (\textit{exact}) or apply the cached approximation (\textit{skip}); the quality-efficiency trade-off is governed entirely by this binary routing policy.

\textbf{Open-Loop Budget Scheduling and Its Limitations.}
Existing schedulers~\cite{liu2024easycache} frame routing as a cumulative risk test: a scalar budget $B_t$ accumulates a heuristic risk increment per timestep, and a skip is taken whenever $B_t < \tau$ for a fixed threshold $\tau$ (full update rule in Appendix~\ref{app:openloop}). This open-loop design has three structural limitations: (i)~a global $\tau$ cannot accommodate heterogeneous dynamics across early, mid, and late phases; (ii)~the absence of a staleness constraint permits unbounded skip runs when inputs change slowly; and (iii)~the budget is never validated against approximation error, so systematic bias propagates uncorrected. AlignCache addresses these limitations through phase-adaptive routing, staleness-aware guarding, and closed-loop threshold calibration~(\S\ref{sec:method}).

\section{Methodology: AlignCache}
\label{sec:method}

\begin{figure}[t]
    \centering
    \includegraphics[width=\textwidth]{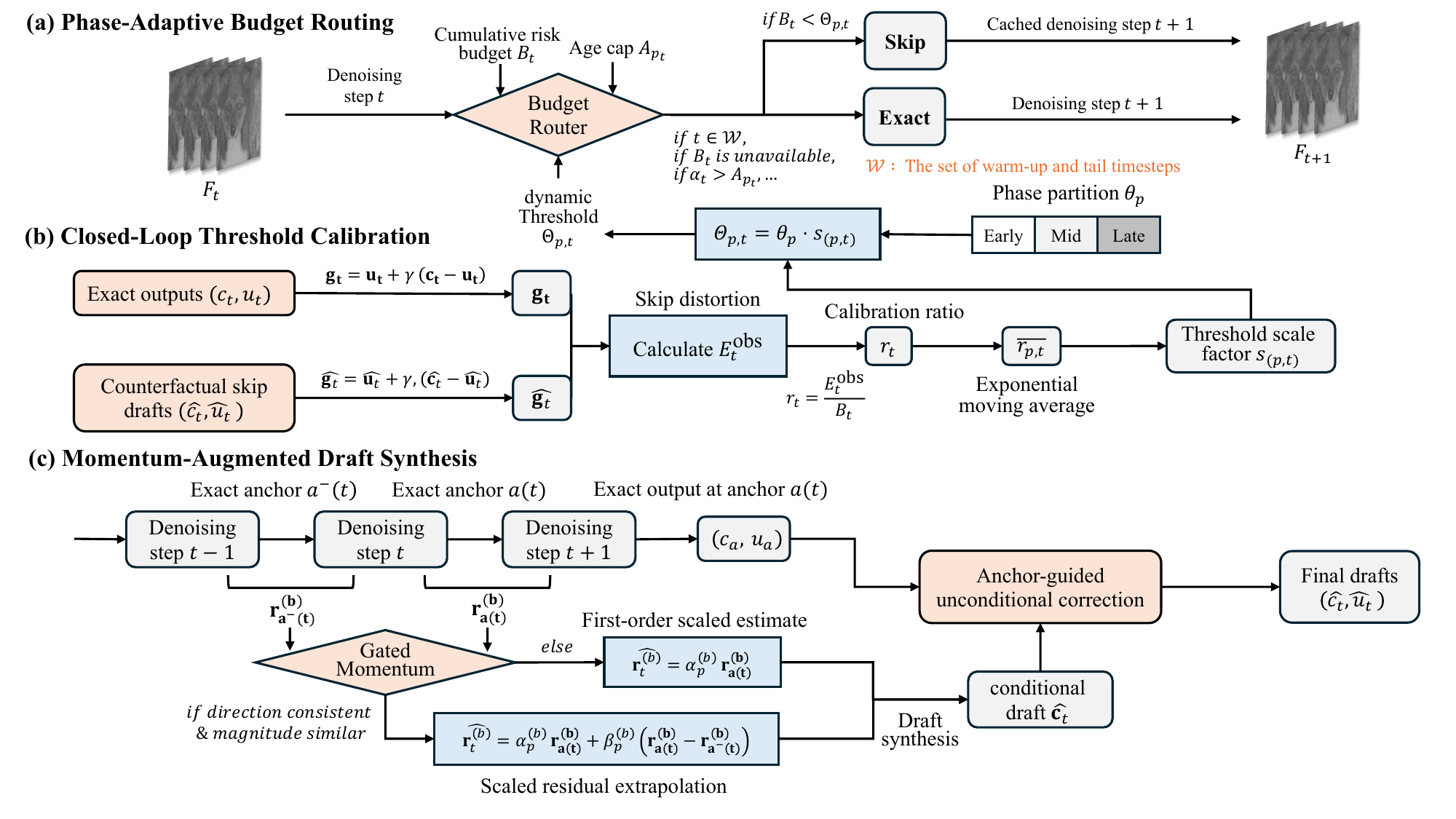}
    \caption{\textbf{Overview of AlignCache.} 
\textbf{(a) Pipeline of phase-adaptive budget routing.} The system decides whether to skip or perform exact computation based on the cumulative risk budget \(B_t\), age cap \(A_{p_t}\), and dynamic threshold \(\theta_{p,t}\). \(F_t\) is the input at timestep \(t\), \(\alpha_t\) the anchor age, and \(\mathcal{W}\) denotes warm-up and tail timesteps unconditionally computed exactly.
\textbf{(b) Workflow of closed-loop threshold calibration.} The process computes exact outputs and counterfactual skip drafts, then derives the skip distortion. The calibration ratio is updated from observed distortion and combined with the dynamic threshold. The resulting scale factor and phase partition \(\theta_p\) determine \(\theta_{p,t}\) in part (a).
\textbf{(c) Process of momentum-augmented draft synthesis.} The current residual is predicted from the two most recent exact anchors \(a(t)\) and \(a^{-}(t)\), applying momentum correction when direction and magnitude align, or a first-order estimate otherwise. The final drafts \((\hat{\mathbf{c}}_t, \hat{\mathbf{u}}_t)\) feed into part (b).
    }
    \label{fig:AlignCache_fig2}
\end{figure}


We consider CFG sampling over $T$ denoising timesteps. At each timestep $t$, the model evaluates conditional and unconditional forward passes, with conditional input $\mathbf{x}_t$, exact conditional output $\mathbf{y}_t$, and CFG-guided output $\mathbf{g}_t$. AlignCache decides at each $t$ whether to execute a full forward pass (\textit{exact}) or reuse cached computations (\textit{skip}). As illustrated in Figure~\ref{fig:AlignCache_fig2}, our framework comprises three coupled components: (i)~a \emph{phase-adaptive budget routing} mechanism that assigns phase-specific thresholds and staleness constraints to segments of the denoising trajectory~(\S\ref{sec:routing}); (ii)~a \emph{closed-loop threshold calibration} module that leverages counterfactual observations from exact computations to adjust decision boundaries~(\S\ref{sec:calibration}); and (iii)~a \emph{momentum-augmented draft synthesis} strategy that improves skip-mode fidelity, enabling more aggressive caching with quality guarantees~(\S\ref{sec:draft}). Unlike prior open-loop schedulers comparing a cumulative budget against a fixed threshold $\tau$, AlignCache replaces this static decision boundary with a dynamic, phase-aware, and self-correcting threshold $\Theta_{p,t}$, transforming scheduling from a risk accumulator into a feedback-driven closed-loop controller.

\subsection{Phase-Adaptive Budget Routing}
\label{sec:routing}

We partition the denoising trajectory into three structurally distinct phases, \texttt{early}, \texttt{mid}, and \texttt{late}, based on transition ratios $(\rho_{\mathrm{e}},\, \rho_{\mathrm{l}})$ over the normalized progress $\rho_t = t / \max(T{-}1, 1)$, with each phase $p$ governed by an independent base threshold $\theta_p$ that permits conservative policies in structurally sensitive phases and aggressive ones in stable phases. We maintain a cumulative risk budget $B_t = B_{t-1} + \Delta B_t$ that accrues a normalized input perturbation amplified by an adaptive motion gain $g_{a(t)}$ since the last exact computation, and resets upon each exact step (full update rules including the motion gain in Appendix~\ref{app:routing}). To prevent unbounded skip runs when inputs change slowly, we further introduce a staleness guard: with anchor age $\alpha_t = t - t_{\mathrm{last}}$ and per-phase cap $A_p$, exact computation is forced whenever $\alpha_t > A_{p_t}$. The unified routing rule is:
\begin{equation}
\label{eq:route}
\mathrm{Route}(t) =
\begin{cases}
\textit{exact}, & \text{if } t \in \mathcal{W} \text{ or } B_t \text{ unavailable or } \alpha_t > A_{p_t}, \\[2pt]
\textit{skip},  & \text{if } B_t < \Theta_{p,t}, \\[2pt]
\textit{exact}, & \text{otherwise},
\end{cases}
\end{equation}
where $\mathcal{W}$ is the set of warm-up and tail timesteps that are unconditionally computed exactly, and the \emph{dynamic threshold} is $\Theta_{p,t} = \theta_p \cdot s_{p,t}$. The base $\theta_p$ encodes phase-specific conservativeness, while the online scale factor $s_{p,t}$ is maintained by the closed-loop calibration module described next, enabling the scheduler to self-adjust according to the evolving fidelity of its own predictions.

\subsection{Closed-Loop Threshold Calibration}
\label{sec:calibration}

A limitation of open-loop schedulers is that the budget $B_t$ may systematically deviate from the true approximation error, causing excessive skipping or unnecessary exact computations. We close this gap by exploiting each budget-triggered exact computation as a calibration opportunity.

\textbf{Counterfactual Error Measurement.}
When the scheduler triggers an exact step because $B_t \ge \Theta_{p_t,t}$, we first construct counterfactual skip drafts $(\hat{\mathbf{c}}_t,\, \hat{\mathbf{u}}_t)$ via the draft synthesis module~(\S\ref{sec:draft}) \emph{before} the true forward pass. After obtaining the exact outputs $(\mathbf{c}_t,\, \mathbf{u}_t)$, we form their guided compositions $\mathbf{g}_t = \mathbf{u}_t + \gamma(\mathbf{c}_t - \mathbf{u}_t)$ and $\hat{\mathbf{g}}_t = \hat{\mathbf{u}}_t + \gamma(\hat{\mathbf{c}}_t - \hat{\mathbf{u}}_t)$, and measure the observed error at the guided level via relative MSE:
\begin{equation}
\label{eq:obs_error}
E_t^{\mathrm{obs}} = \frac{\|\hat{\mathbf{g}}_t - \mathbf{g}_t\|_2^2}{\|\mathbf{g}_t\|_2^2}.
\end{equation}
This directly quantifies the distortion the skip approximation would have injected into the final sample, which is more informative than per-branch comparison since $\mathbf{g}_t$ is the quantity that ultimately determines generation quality.

\textbf{Online Scale Update with Selective Feedback.}
We define the calibration ratio $r_t = E_t^{\mathrm{obs}} / B_t$, where $r_t > 1$ indicates that the budget underestimates the true error and $r_t < 1$ signals overestimation. We maintain a clipped per-phase exponential moving average $\bar{r}_{p,t}$ and compute the threshold scale factor $s_{p,t}$ via an inverse-clipping rule (full update equations in Appendix~\ref{app:calibration}). The inverse relationship ensures a self-correcting feedback loop: when $\bar{r}_{p,t} > 1$, $s_{p,t}$ decreases, tightening $\Theta_{p,t}$ and triggering more frequent exact computations; when $\bar{r}_{p,t} < 1$, the threshold relaxes for additional skips. To avoid contaminating the calibration signal with structural scheduling events, we restrict feedback updates to exact computations triggered \emph{solely} by the budget criterion ($B_t \ge \Theta_{p,t}$), excluding warm-up, bootstrap, and staleness-triggered events.

\subsection{Momentum-Augmented Draft Synthesis}
\label{sec:draft}

The fidelity of skip-mode outputs affects both generation quality and the reliability of the closed-loop calibration signal. We design a draft synthesis module that goes beyond naive residual reuse via direction-magnitude gated extrapolation and anchor-guided correction.

\textbf{Scaled Residual Extrapolation with Gated Momentum.}
For each branch $b \in \{\mathrm{cond}, \mathrm{uncond}\}$, given the residuals at the two most recent exact anchors $a(t)$ and $a^{-}(t)$, we predict the current residual via a first-order scaled estimate augmented with a gated second-order momentum correction:
\begin{equation}
\label{eq:predictor}
\hat{\mathbf{r}}_t^{(b)} = \alpha_p^{(b)}\, \mathbf{r}_{a(t)}^{(b)} + \beta_p^{(b)}\!\left(\mathbf{r}_{a(t)}^{(b)} - \mathbf{r}_{a^{-}(t)}^{(b)}\right),
\end{equation}
where $\alpha_p^{(b)}$ is a phase-dependent scaling coefficient and $\beta_p^{(b)}$ is an adaptive momentum weight modulated by a direction-magnitude gating function (full form in Appendix~\ref{app:draft}). The direction gate suppresses extrapolation when consecutive anchor residuals diverge in orientation, while the magnitude gate attenuates the correction when residual norms change rapidly, gracefully decaying $\beta_p^{(b)}$ toward zero during transient dynamics. Compared with static residual reuse ($\hat{\mathbf{y}}_t^{(b)} = \mathbf{x}_t^{(b)} + \mathbf{r}_{a(t)}^{(b)}$), this extrapolation captures the evolving trend of residuals and reduces drift during extended skip runs.

\textbf{Anchor-Guided Unconditional Correction.}
We further refine the unconditional draft by exploiting the structural correlation between conditional and unconditional branches at the most recent anchor. Let $(\mathbf{c}_a,\, \mathbf{u}_a)$ denote the exact outputs at anchor $a(t)$ and $\hat{\mathbf{c}}_t$ the current conditional draft. The final unconditional draft blends the extrapolation baseline $\hat{\mathbf{u}}_t^{\mathrm{base}}$ with an anchor-transferred estimate $\tilde{\mathbf{u}}_t = \mathbf{u}_a + \alpha_p (\hat{\mathbf{c}}_t - \mathbf{c}_a)$ via a blending weight $\omega$ (Appendix~\ref{app:draft}). This leverages the cond-uncond coupling inherent in CFG to produce more faithful unconditional drafts, yielding more accurate guided approximations during both skip execution and counterfactual error measurement.

\begin{table}[t]
\centering
\caption{Comparison between other acceleration strategies and our AlignCache. We report inference latency, speedup, visual retention metrics (PSNR, SSIM, LPIPS, and GMSD), and VBench Total score on representative video generation models. The baseline is
reported with the original $T$ denoising steps. ``$+$'' denotes applying the corresponding acceleration strategy on top of the baseline.}
\label{tab:main_results}
\setlength{\tabcolsep}{6pt}
\renewcommand{\arraystretch}{1.1}
\resizebox{\linewidth}{!}{%
\begin{tabular}{l cc cccc c}
\toprule
\multirow{2}{*}{\textbf{Methods}} &
\multicolumn{2}{c}{\textbf{Efficiency}} &
\multicolumn{4}{c}{\textbf{Visual Retention}} &
\textbf{VBench} \\
\cmidrule(lr){2-3} \cmidrule(lr){4-7} \cmidrule(lr){8-8}
 & Latency (s)~$\downarrow$ & Speedup~$\uparrow$ &
PSNR~$\uparrow$ & SSIM~$\uparrow$ & LPIPS~$\downarrow$ & GMSD~$\downarrow$ &
Total (\%)~$\uparrow$ \\
\midrule
\multicolumn{8}{c}{\textit{Wan2.1-1.3B~\cite{wang2025wan} (81 frames, 832$\times$480)}} \\
\midrule
Wan2.1-1.3B ($T{=}50$)                           & 131.9 & 1.00$\times$ & -- & -- & -- & -- & 81.64 \\
\quad + PAB~\cite{zhao2024pab}              & 74.2  & 1.78$\times$ & 21.42 & 0.780 & 0.162 & 0.164 & 81.00 \\
\quad + TeaCache~\cite{liu2025timestep}     & 86.2  & 1.53$\times$ & 27.57 & 0.903 & 0.055 & 0.100 & 81.23 \\
\quad + FasterCache~\cite{lv2024fastercache}& 75.9  & 1.74$\times$ & 26.35 & 0.891 & 0.063 & 0.109 & 81.41 \\
\quad + EasyCache~\cite{liu2024easycache}   & 61.1  & 2.19$\times$ & 28.38 & 0.903 & 0.051 & 0.089 & 81.04 \\
\rowcolor{ourrow}
\quad + \textbf{AlignCache (Ours)}          & \textbf{57.5} & \textbf{2.29$\times$} & \textbf{29.80} & \textbf{0.925} & \textbf{0.038} & \textbf{0.080} & 81.28 \\
\midrule
\multicolumn{8}{c}{\textit{Open-Sora-1.2~\cite{zheng2024open} (102 frames, 848$\times$480)}} \\
\midrule
Open-Sora ($T{=}30$)                        & 70.2 & 1.00$\times$ & -- & -- & -- & -- & 80.07 \\
\quad + PAB~\cite{zhao2024pab}              & 48.3 & 1.45$\times$ & 18.76 & 0.716 & 0.319 & 0.188 & 75.62 \\
\quad + TeaCache~\cite{liu2025timestep}     & 47.2 & 1.49$\times$ & 22.10 & 0.802 & 0.179 & 0.156 & 79.61 \\
\quad + FasterCache~\cite{lv2024fastercache}& 45.5 & 1.54$\times$ & 25.29 & 0.861 & 0.115 & 0.124 & 80.08 \\
\quad + EasyCache~\cite{liu2024easycache}   & 41.1 & 1.75$\times$ & 25.87 & 0.876 & 0.112 & 0.114 & 79.58 \\
\rowcolor{ourrow}
\quad + \textbf{AlignCache (Ours)}          & \textbf{38.1} & \textbf{1.86$\times$} & \textbf{26.06} & \textbf{0.878} & \textbf{0.106} & \textbf{0.114} & 79.59 \\
\midrule
\multicolumn{8}{c}{\textit{HunyuanVideo~\cite{kong2024hunyuanvideo} (65 frames, 864$\times$480)}} \\
\midrule
HunyuanVideo ($T{=}50$)                     & 208.7 & 1.00$\times$ & -- & -- & -- & -- & 82.55 \\
\quad + PAB~\cite{zhao2024pab}              & 133.4 & 1.56$\times$ & 32.36 & 0.939 & 0.050 & 0.061 & 82.17 \\
\quad + TeaCache~\cite{liu2025timestep}     & 98.7  & 2.11$\times$ & 22.52 & 0.788 & 0.206 & 0.150 & 82.42 \\
\quad + FasterCache~\cite{lv2024fastercache}& 130.1 & 1.60$\times$ & 27.13 & 0.874 & 0.119 & 0.109 & 82.64 \\
\quad + EasyCache~\cite{liu2024easycache}   & 101.9 & 2.09$\times$ & 32.33 & 0.923 & 0.063 & 0.059 & 82.08 \\
\rowcolor{ourrow}
\quad + \textbf{AlignCache (Ours)}          & \textbf{98.1} & \textbf{2.15$\times$} & \textbf{34.23} & \textbf{0.945} & \textbf{0.046} & \textbf{0.049} & 82.16 \\
\bottomrule
\end{tabular}}
\end{table}

\section{Experiments and Results}

\subsection{Setup}

\textbf{Baselines.} We compare AlignCache against state-of-the-art training-free cache-based accelerators, including PAB~\cite{zhao2024pab}, TeaCache~\cite{liu2025timestep}, FasterCache~\cite{lv2024fastercache}, and EasyCache~\cite{liu2024easycache}, using their official implementations whenever available, and faithfully re-implementing them on backbones not directly supported by the original code.

\textbf{Models, Metrics, and Datasets.} We evaluate on three DiT-based video generators: Wan2.1-1.3B~\cite{wang2025wan} ($81$ frames, $832{\times}480$, $T{=}50$), Open-Sora-1.2~\cite{zheng2024open} ($102$ frames, $848{\times}480$, $T{=}30$), and HunyuanVideo~\cite{kong2024hunyuanvideo} ($65$ frames, $864{\times}480$, $T{=}50$). For each model, we measure end-to-end latency and speedup, reference-fidelity metrics against the unaccelerated output (PSNR, SSIM, LPIPS, GMSD), and the VBench~\cite{huang2024vbench} Total score on a subset of official prompts with $5$ or $10$ samples per prompt.

\textbf{Implementation Details.} Hyperparameters are tuned by sweeping $5$ prompts $\times\,2$ seeds. For Wan2.1-1.3B, the base threshold is $0.065$ with $12$ warm-up exact CFG pairs and phase scales $(1.15,\,1.00,\,0.80)$ for early/mid/late. For Open-Sora-1.2 and HunyuanVideo, the base thresholds are $0.11$ and $0.024$, with $8$ and $10$ warm-up pairs, respectively. Phase boundaries are $(15\%,\,86\%)$ for Wan2.1-1.3B/Open-Sora-1.2 and $(18\%,\,72\%)$ for HunyuanVideo. Main results are obtained on NVIDIA RTX PRO 6000-96GB; ablations on Wan2.1-1.3B use $5$ prompts $\times\,2$ seeds (42, 954) on a single NVIDIA H100-80GB.

\subsection{Main Results}

Table~\ref{tab:main_results} compares AlignCache with prior cache-based accelerators across the three backbones. AlignCache simultaneously attains the \emph{highest speedup} and the \emph{best fidelity} on every model, dominating the Pareto frontier without exception.

\begin{wraptable}[24]{r}{0.5\textwidth}
\vspace{-12pt}
\centering
\caption{Ablations on \textbf{scheduler-related} axes on Wan2.1-1.3B. Representative configurations shown; full grid in App.~\ref{app:full-ablation}.}
\label{tab:abl_sched}
\setlength{\tabcolsep}{3pt}
\renewcommand{\arraystretch}{0.9}
\scriptsize
\begin{tabular}{l cc ccc}
\toprule
Variant & Lat.(s)$\downarrow$ & Spd.$\uparrow$ & PSNR$\uparrow$ & SSIM$\uparrow$ & LPIPS$\downarrow$ \\
\midrule
\rowcolor{gray!12}
\multicolumn{6}{c}{\textit{(a) Warm-up steps $W$}} \\
\midrule
$W{=}5$  & 42.73 & 2.29$\times$ & 23.68 & 0.829 & 0.123 \\
$W{=}7$  & 43.12 & 2.26$\times$ & 24.40 & 0.850 & 0.110 \\
$W{=}9$  & 43.72 & 2.22$\times$ & 24.51 & 0.862 & 0.104 \\
$W{=}12$ & 46.82 & 2.06$\times$ & 25.97 & 0.882 & 0.086 \\
$W{=}14$ & 49.03 & 1.97$\times$ & 26.34 & 0.887 & 0.078 \\
\midrule
\rowcolor{gray!12}
\multicolumn{6}{c}{\textit{(b) Error threshold $\theta_p$}} \\
\midrule
$0.0325$ & 53.77 & 1.81$\times$ & 27.20 & 0.904 & 0.070 \\
$0.0400$ & 50.42 & 1.91$\times$ & 25.48 & 0.870 & 0.090 \\
$0.0425$ & 47.21 & 2.06$\times$ & 25.58 & 0.880 & 0.088 \\
$0.0600$ & 41.75 & 2.31$\times$ & 24.05 & 0.843 & 0.114 \\
$0.0700$ & 39.35 & 2.45$\times$ & 23.65 & 0.832 & 0.123 \\
\midrule
\rowcolor{gray!12}
\multicolumn{6}{c}{\textit{(c) Phase ratio $(\rho_e,\rho_l)$}} \\
\midrule
$(0.125,0.88)$ & 44.29 & 2.19$\times$ & 25.11 & 0.873 & 0.095 \\
$(0.150,0.85)$ & 45.12 & 2.14$\times$ & 24.61 & 0.859 & 0.102 \\
$(0.150,0.80)$ & 45.15 & 2.14$\times$ & 24.61 & 0.859 & 0.102 \\
$(0.175,0.88)$ & 44.28 & 2.19$\times$ & 25.05 & 0.873 & 0.096 \\
$(0.200,0.86)$ & 45.14 & 2.14$\times$ & 24.65 & 0.859 & 0.100 \\
\bottomrule
\end{tabular}
\vspace{-8pt}
\end{wraptable}

\textbf{Pareto dominance across backbones.} On Wan2.1-1.3B, AlignCache reaches $2.29\times$ speedup with $29.80$\,dB PSNR and $0.038$ LPIPS, improving over the strongest baseline EasyCache by $+1.42$\,dB PSNR and $-25\%$ LPIPS while being faster ($2.29\times$ vs.\ $2.19\times$). On HunyuanVideo, the margin widens to $+1.90$\,dB PSNR and $-22\%$ LPIPS at a higher $2.15\times$ speedup. On Open-Sora-1.2, AlignCache delivers the largest speedup ($1.86\times$) among all methods while still leading on PSNR, SSIM, and LPIPS. The consistency of this gain across three architectures with different sampling lengths and aspect ratios indicates that the improvement stems from a model-agnostic property of the scheduler rather than backbone-specific tuning.

\textbf{Source of the joint speed-quality improvement.} A natural concern is whether AlignCache simply tightens the threshold to trade speed for quality. The data argue otherwise: on Wan2.1-1.3B, AlignCache simultaneously \emph{reduces} latency relative to EasyCache ($57.5$\,s vs.\ $61.1$\,s) \emph{and} improves PSNR by over a full dB. Such co-improvement is enabled by two complementary mechanisms acting in concert. The momentum-augmented draft synthesis produces higher-fidelity skip outputs by capturing residual curvature beyond static reuse, directly reducing approximation error along extended skip runs and accounting for a substantial portion of the PSNR gain. The closed-loop calibration further redirects the scheduler toward exact computation when the calibration ratio $\bar{r}_{p,t}$ exceeds $1$, ensuring that residual budget is spent on the most informative timesteps. Inspecting the per-step routing logs, we observe that AlignCache skips aggressively in the middle phase where residuals are smooth and reverts to exact computation in the late phase when feedback signals indicate growing distortion, behavior that an open-loop scheduler with a single global $\tau$ structurally cannot reproduce.

\textbf{VBench scores reflect a coarser axis than visual fidelity.} VBench Total is comparable across methods because reference-free aggregate scores conflate true generation quality with prompt-induced variation; the more discriminative signal lies in PSNR/LPIPS against the unaccelerated baseline, where AlignCache consistently leads by a clear margin. AlignCache attains $0.8128/0.7959/0.8216$ on the three backbones, on par with the best baseline on each, indicating that the closed-loop calibration recovers the original generation distribution rather than over-smoothing it into a more ``VBench-friendly'' but distorted output.

\subsection{Ablation Studies}
\label{sec:ablation}

To understand the contribution of each design choice in AlignCache, we conduct controlled ablations along five orthogonal axes on \textit{Wan2.1-1.3B}~\cite{wang2025wan}, organized into two conceptual layers: \emph{scheduler-related} axes (warm-up steps, error threshold, phase ratio) in Table~\ref{tab:abl_sched} and \emph{predictor-related} axes (motion gain, residual extrapolation order) in Table~\ref{tab:abl_pred}. All variants use $5$ prompts $\times$ $2$ random seeds against an unaccelerated baseline whose mean latency is $96.54$\,s on a single NVIDIA H100-80GB, and we report end-to-end latency, sampling time speedup, and reference-fidelity (PSNR/SSIM/LPIPS) measured against the baseline output.

\begin{wraptable}[18]{r}{0.5\textwidth}
\vspace{-12pt}
\centering
\caption{Ablations on \textbf{predictor-related} axes on Wan2.1-1.3B. Subgroups (a)-(b) vary motion-gain parameters and residual extrapolation order, respectively.}
\label{tab:abl_pred}
\setlength{\tabcolsep}{3pt}
\renewcommand{\arraystretch}{0.85}
\scriptsize
\begin{tabular}{l cc ccc}
\toprule
Variant & Lat.(s)$\downarrow$ & Spd.$\uparrow$ & PSNR$\uparrow$ & SSIM$\uparrow$ & LPIPS$\downarrow$ \\
\midrule
\rowcolor{gray!12}
\multicolumn{6}{c}{\textit{(a) Motion gain $(\beta_0^{(p)}, s)$}} \\
\midrule
$(0.50,\,1.00)$ & 49.76 & 1.94$\times$ & 25.58 & 0.873 & 0.089 \\
$(0.75,\,1.00)$ & 62.23 & 1.56$\times$ & 30.21 & 0.937 & 0.042 \\
$(0.80,\,0.75)$ & 63.73 & 1.51$\times$ & 29.23 & 0.930 & 0.045 \\
$(0.80,\,0.95)$ & 67.25 & 1.44$\times$ & 32.70 & 0.954 & 0.030 \\
$(0.80,\,1.00)$ & 68.23 & 1.41$\times$ & 32.22 & 0.954 & 0.029 \\
$(0.80,\,1.05)$ & 69.38 & 1.39$\times$ & 33.69 & 0.959 & 0.026 \\
$(0.80,\,1.15)$ & 71.15 & 1.36$\times$ & 34.35 & 0.963 & 0.022 \\
$(0.80,\,1.25)$ & 73.37 & 1.31$\times$ & 33.49 & 0.961 & 0.024 \\
$(0.85,\,1.00)$ & 79.70 & 1.21$\times$ & 36.69 & 0.973 & 0.015 \\
$(0.90,\,1.00)$ & 96.42 & 1.00$\times$ & --    & 1.000 & 0.000 \\
\midrule
\rowcolor{gray!12}
\multicolumn{6}{c}{\textit{(b) Residual extrapolation order}} \\
\midrule
None             & 43.45 & 2.23$\times$ & 24.51 & 0.843 & 0.118 \\
1st-order only   & 44.13 & 2.19$\times$ & 24.04 & 0.841 & 0.117 \\
1st\,+\,2nd      & 45.12 & 2.14$\times$ & 24.61 & 0.859 & 0.102 \\
\bottomrule
\end{tabular}
\vspace{-8pt}
\end{wraptable}

\textbf{Effect of warm-up steps.} Subgroup (a) of Table~\ref{tab:abl_sched} shows that increasing the warm-up budget from $W{=}5$ to $W{=}14$ raises PSNR by $+2.66$\,dB and reduces LPIPS by $37\%$, while the speedup decreases monotonically from $2.29\times$ to $1.97\times$. This trade-off arises because warm-up serves as a statistical prerequisite for closed-loop calibration: with too few warm-up steps, the EMA in Eq.~\eqref{eq:ema} (Appendix~\ref{app:calibration}) lacks sufficient samples to stabilize, and the resulting noisy feedback degrades routing decisions, which is why $W{=}5$ achieves a similar speedup to $W{=}9$ but $0.83$\,dB lower PSNR. Beyond a sufficient warm-up budget, additional exact steps forfeit skip opportunities without commensurate quality gain, indicating that warm-up duration governs a calibration-stability regime that grows with $W$.

\textbf{Effect of error base threshold.} Subgroup (b) shows that each $0.0025$ increment in $\theta_p$ buys roughly $+0.05\times$ to $+0.10\times$ extra speedup with a graceful degradation in fidelity across the swept range. The fidelity drop accelerates at higher thresholds when skip runs grow long enough that the residual predictor alone cannot fully compensate, exposing a tight coupling between threshold and predictor: raising $\theta_p$ without strengthening the predictor amplifies extrapolation error super-linearly along skip runs, so the achievable quality-speed trade-off is shaped jointly by these two layers rather than by the threshold alone.

\textbf{Effect of phase ratio.} Subgroup (c) indicates that the phase ratio is the \emph{least sensitive} axis: PSNR varies by less than $0.55$\,dB and SSIM by less than $0.015$ across the full $11$-point grid. This insensitivity is itself evidence that the closed-loop design is doing what it claims: the calibrated $\Theta_{p,t}$ absorbs small misspecifications in phase boundaries through online adjustment of $s_{p,t}$, shifting the dominant performance bottleneck to threshold and motion-gain settings rather than partitioning. The flatness of the grid further confirms that the inductive split between transient and converged dynamics is qualitative rather than knife-edge, so phase boundaries can be treated as a structural prior rather than a tightly tuned hyperparameter.

\textbf{Effect of motion gain.} Subgroup (a) of Table~\ref{tab:abl_pred} reveals that motion gain is the strongest \emph{quality-recovery} knob: increasing $\beta_0^{(p)}$ from $0.50$ to $0.85$ lifts PSNR by $+11.1$\,dB and reduces LPIPS by $83\%$, while $\beta_0^{(p)}{=}0.90$ collapses the speedup to $1.00\times$ as the gain becomes so conservative that the controller never skips. The order-of-magnitude swing in PSNR with a small change in $\beta_0^{(p)}$ indicates that skip-mode fidelity is governed by the curvature of the residual trajectory rather than its raw magnitude, and the second-order momentum term, when correctly gated, recovers most of the lost signal that anchor reuse cannot reconstruct. The threshold scale $s$ provides finer control: holding $\beta_0^{(p)}{=}0.80$ and varying $s$ from $0.75$ to $1.25$ trades $+5.1$\,dB PSNR for $-0.20\times$ speedup, giving practitioners a smooth quality--efficiency dial that can be adjusted along the trade-off curve.

\textbf{Effect of residual extrapolation.} Subgroup (b) shows that disabling residual extrapolation entirely costs $0.10$\,dB in PSNR and $16\%$ in LPIPS for only $+0.09\times$ extra speedup, and the first-order-only variant likewise underperforms the full predictor. The second-order momentum correction in Eq.~\eqref{eq:predictor} therefore plays a dual role: it captures the curvature of the residual trajectory during extended skip runs to improve skip-mode fidelity, and it sharpens the calibration signal $E_t^{\mathrm{obs}}$, since a more accurate counterfactual draft yields a less biased estimate of skip distortion. Removing it therefore degrades both generation quality and downstream calibration accuracy, compounding the loss across both layers of the framework.

\textbf{Cross-platform efficiency.} The $2.29\times$ speedup on Wan2.1-1.3B transfers consistently across NVIDIA Hopper and Blackwell architectures, reaching $42.4$\,s end-to-end latency on H100-80GB and $57.5$\,s on RTX PRO 6000-96GB, both at the same $2.29\times$ speedup ratio over the unaccelerated baseline. The identical speedup factor across two distinct architectures demonstrates that the acceleration is computation-bound rather than memory-bound and that the scheduling overhead is negligible.

\subsection{Case Studies}
\label{sec:case_studies}

\begin{figure}[htbp]
    \centering
    \includegraphics[width=\textwidth]{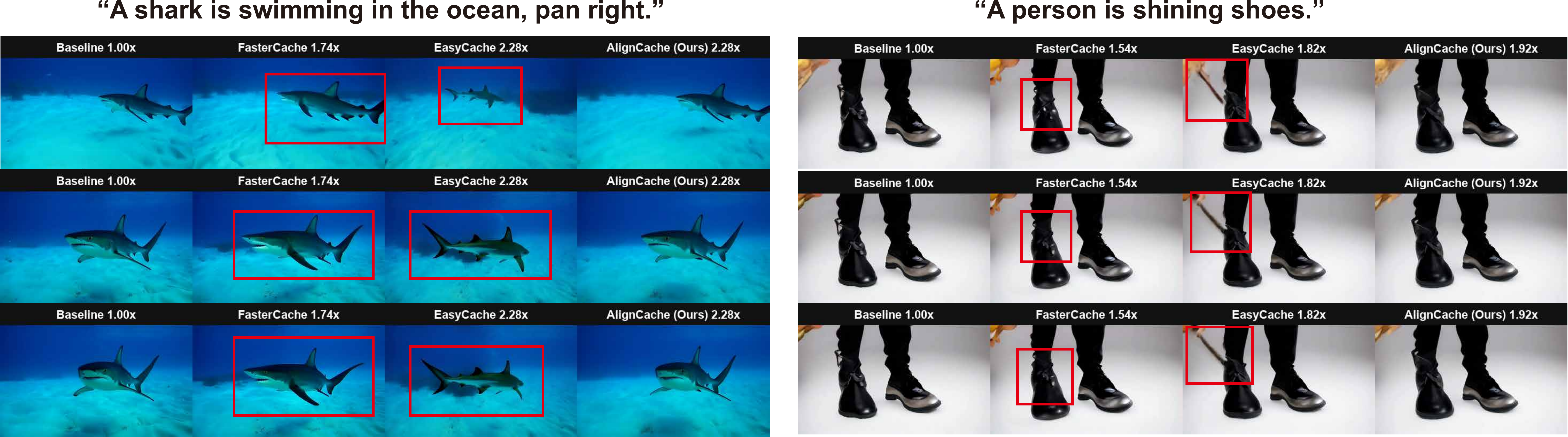}
    \caption{\textbf{Qualitative comparison} on \emph{``A shark is swimming in the ocean, pan right.''} (left, Wan2.1-1.3B) and \emph{``A person is shining shoes.''} (right, Open-Sora-1.2) under matched or higher speedup settings. Red boxes highlight regions where competing methods introduce noticeable distortions, while AlignCache faithfully preserves object geometry and fine details at a comparable or higher speedup. More test method case studies can be found in the Appendix~\ref{app:case_studies}.}
    \label{fig:case_study}
\end{figure}

Figure~\ref{fig:case_study} presents qualitative comparisons on representative VBench prompts spanning diverse motion patterns. On \emph{``A shark is swimming in the ocean, pan right.''} (Wan2.1-1.3B), FasterCache and EasyCache visibly distort the shark's silhouette and fin geometry under their respective caching policies, while AlignCache preserves the object boundary and motion direction at a comparable or higher speedup without temporal flicker. On \emph{``A person is shining shoes.''} (Open-Sora-1.2), competing methods introduce blurring artifacts and texture inconsistencies on the shoe surface and hand region (red boxes), whereas AlignCache faithfully retains fine leather textures and high-frequency edges throughout the entire sequence. The qualitative gap concentrates on semantically salient regions, object boundaries, faces, and high-frequency textures, precisely where direct residual reuse incurs the largest approximation error in the late denoising phase. Through $E_t^{\mathrm{obs}}$, the closed-loop calibrator detects these high-error events and adaptively forces exact computations, preventing the open-loop failure mode of accumulating distortion past the perceptual threshold and explaining why the LPIPS gap exceeds the PSNR gap on average across prompts. 



\section{Conclusion and Future Works}

\textbf{Conclusion.} We presented \textbf{AlignCache}, a closed-loop caching framework that addresses three structural limitations of existing adaptive schedulers: heuristic criteria misaligned with generation quality, single global thresholds insensitive to phase dynamics, and uncorrected systematic bias in budget estimation. By coupling phase-adaptive budget routing, counterfactual error measurement at the CFG-guided output level, and momentum-augmented draft synthesis, AlignCache transforms the scheduling paradigm from a static risk accumulator into a feedback-driven, self-correcting controller. Experiments on Wan2.1-1.3B, Open-Sora-1.2, and HunyuanVideo show that AlignCache consistently achieves the best speed-quality Pareto frontier, simultaneously delivering higher speedup and better fidelity than prior cache-based accelerators.

\textbf{Future Works.} Several directions remain open. First, the closed-loop principle behind AlignCache is not specific to caching, the same counterfactual error measurement could calibrate other training-free accelerators such as token pruning or sparse attention. Second, current calibration treats each prompt independently; learning a prompt-conditioned prior over $\bar{r}_{p,t}$ could further reduce warm-up overhead and unlock higher speedups on short trajectories. Third, combining AlignCache with orthogonal acceleration techniques such as quantization, distillation, or compiler-level kernel fusion could yield compounded speedups while the closed-loop controller absorbs the additional approximation error introduced by these layers and preserves stable quality across heterogeneous deployment hardware. Finally, extending AlignCache to long-form video generation and multi-modal diffusion pipelines, where redundancy patterns differ substantially from text-to-video, is a natural next step toward a unified, self-correcting acceleration framework for generative models at scale.

\bibliographystyle{plain}
\bibliography{main}








\appendix

\section{Technical Appendices and Supplementary Material}

\subsection{Implementation Details}
\label{app:impl}

Due to the modular design of AlignCache, the method has several hyperparameters. To achieve a better quality-speed trade-off, we tune these hyperparameters by sweeping over 5 prompts with 2 seeds per prompt. The final hyperparameters are as follows. For Wan2.1-1.3B, we set the base cache threshold to 0.065 with 12 warm-up exact CFG pairs, and use phase-adaptive threshold scales of 1.15, 1.00, and 0.80 for the early, middle, and late denoising phases, respectively, corresponding to phase thresholds of 0.07475, 0.065, and 0.052. For Open-Sora-1.2, we use a base threshold of 0.11, 8 warm-up pairs, and phase thresholds of 0.1045, 0.1265, and 0.11. For HunyuanVideo, we use the setting with a base threshold of 0.024, 10 warm-up pairs, and phase thresholds of 0.0252, 0.0288, and 0.0324. Unless otherwise specified, the phase boundaries are set to 15\% and 86\% of the denoising trajectory for Wan2.1-1.3B/Open-Sora-1.2, and 18\% and 72\% for HunyuanVideo. For the principal quantitative evaluations reported on a subset of the VBench default prompts, we generate 5 or 10 video samples, depending on the official recommendations for different evaluation dimensions, with different seeds for each prompt on NVIDIA RTX PRO 6000-96GB GPUs. To expedite ablation studies, we generate 2 videos per prompt with seeds 42 and 954, using Wan2.1-1.3B as the baseline.

\subsection{Detailed Open-Loop Budget Scheduling}
\label{app:openloop}

In existing open-loop schedulers~\cite{liu2024easycache} referenced in Section~\ref{sec:pre}, the cumulative risk budget $B_t$ is maintained by accumulating, at each timestep, a heuristic risk increment:
\begin{equation}
\label{eq:openloop_budget}
\Delta B_t = g_{a(t)} \cdot \frac{\mathrm{mean}(|\mathbf{x}_t - \mathbf{x}_{t-1}|)}{\mathrm{mean}(|\mathbf{y}_{a(t)}|)},
\end{equation}
where $g_{a(t)}$ is a gain factor estimated from the two most recent anchors. The timestep is skipped whenever $B_t < \tau$ for a fixed global threshold $\tau$, and $B_t$ is reset upon each exact computation. The three structural limitations of this design (single global threshold, lack of staleness constraint, lack of error validation) are discussed in the main text.

\subsection{Detailed Phase-Adaptive Budget Routing}
\label{app:routing}

This section provides the detailed update rules underlying the phase-adaptive routing module summarized in Section~\ref{sec:routing}.

\paragraph{Phase Partition.}
Let $\rho_t = t / \max(T{-}1,\, 1)$ denote the normalized progress at timestep $t$. The phase assignment is:
\begin{equation}
\label{eq:phase}
p_t =
\begin{cases}
\texttt{early}, & \text{if } \rho_t < \rho_{\mathrm{e}}, \\[2pt]
\texttt{mid},   & \text{if } \rho_{\mathrm{e}} \le \rho_t < \rho_{\mathrm{l}}, \\[2pt]
\texttt{late},  & \text{if } \rho_t \ge \rho_{\mathrm{l}},
\end{cases}
\end{equation}
where $\rho_{\mathrm{e}}$ and $\rho_{\mathrm{l}}$ are the early-to-mid and mid-to-late transition ratios, respectively.

\paragraph{Cumulative Risk Budget Update.}
At timestep $t$, the instantaneous input variation is $\delta_t = \mathrm{mean}(|\mathbf{x}_t - \mathbf{x}_{t-1}|)$, and the budget increment is:
\begin{equation}
\label{eq:budget}
\Delta B_t = g_{a(t)} \cdot \frac{\delta_t}{\mathrm{mean}\!\big(|\mathbf{y}_{a(t)}|\big)},
\end{equation}
where $a(t)$ is the index of the most recent exact anchor, $\mathbf{y}_{a(t)}$ is its cached output, and $g_{a(t)}$ is a motion gain factor reflecting local input-output sensitivity. The budget accumulates as $B_t = B_{t-1} + \Delta B_t$ and is reset to zero after each exact computation. Intuitively, Eq.~\eqref{eq:budget} normalizes the input perturbation by the output magnitude and amplifies it by the learned gain, yielding a unit-free risk estimate.

\paragraph{Motion Gain Update.}
The motion gain $g_{a(t)}$ is updated after each exact computation using the ratio of output variation to input variation between the two most recent exact anchors:
\begin{equation}
\label{eq:gain}
g_{a(t)} = \frac{\mathrm{mean}\!\big(|\mathbf{y}_{a(t)} - \mathbf{y}_{a^{-}(t)}|\big)}{\mathrm{mean}\!\big(|\mathbf{x}_{a(t)} - \mathbf{x}_{a^{-}(t)}|\big)},
\end{equation}
where $a^{-}(t)$ denotes the exact anchor immediately preceding $a(t)$. This ratio adapts the budget sensitivity to the local dynamics of the diffusion trajectory.

\paragraph{Dynamic Threshold.}
The dynamic threshold used in the routing rule (Eq.~\eqref{eq:route}) is:
\begin{equation}
\label{eq:dyn_thresh}
\Theta_{p,t} = \theta_p \cdot s_{p,t},
\end{equation}
where $\theta_p$ is the base phase threshold and $s_{p,t}$ is the online scale factor maintained by the closed-loop calibration module (Appendix~\ref{app:calibration}).

\subsection{Detailed Closed-Loop Threshold Calibration}
\label{app:calibration}

This section provides the detailed update equations underlying the closed-loop calibration module summarized in Section~\ref{sec:calibration}. Building on the calibration ratio $r_t = E_t^{\mathrm{obs}} / B_t$ defined in the main text, we maintain a per-phase clipped exponential moving average:
\begin{equation}
\label{eq:ema}
\bar{r}_{p,t} = \lambda\, \bar{r}_{p,t-1} + (1 - \lambda)\, \mathrm{clip}(r_t,\; r_{\min},\; r_{\max}),
\end{equation}
where $\lambda$ is the EMA decay factor controlling the smoothness of the estimate, and $[r_{\min}, r_{\max}]$ are clipping bounds preventing outlier observations from destabilizing the controller. The threshold scale factor is then:
\begin{equation}
\label{eq:scale}
s_{p,t} = \mathrm{clip}\!\left(\frac{1}{\bar{r}_{p,t}},\;\; s_{\min}^{(p)},\;\; s_{\max}^{(p)}\right),
\end{equation}
where $s_{\min}^{(p)}$ and $s_{\max}^{(p)}$ are per-phase bounds. The inverse relationship in Eq.~\eqref{eq:scale} is what makes the loop self-correcting: when the budget consistently underestimates error ($\bar{r}_{p,t} > 1$), the scale factor decreases, tightening $\Theta_{p,t}$ and triggering more frequent exact computations; conversely, when the budget is overly conservative ($\bar{r}_{p,t} < 1$), the threshold relaxes to permit additional skips.

\subsection{Detailed Momentum-Augmented Draft Synthesis}
\label{app:draft}

This section provides the detailed equations underlying the gated momentum and anchor-guided correction summarized in Section~\ref{sec:draft}.

\paragraph{Direction-Magnitude Gated Momentum.}
The adaptive momentum weight $\beta_p^{(b)}$ in Eq.~\eqref{eq:predictor} is modulated by a gating function that jointly evaluates directional coherence and magnitude consistency:
\begin{equation}
\label{eq:beta_gate}
\beta_p^{(b)} = \beta_0^{(p)} \cdot \underbrace{\max\!\Big(0,\;\mathrm{cossim}\!\big(\mathbf{r}_{a(t)}^{(b)},\, \mathbf{r}_{a^{-}(t)}^{(b)}\big)\Big)}_{\text{direction gate}} \cdot \underbrace{\min\!\left(\frac{\|\mathbf{r}_{a(t)}^{(b)}\|}{\|\mathbf{r}_{a^{-}(t)}^{(b)}\|},\;\frac{\|\mathbf{r}_{a^{-}(t)}^{(b)}\|}{\|\mathbf{r}_{a(t)}^{(b)}\|}\right)}_{\text{magnitude gate}},
\end{equation}
where $\beta_0^{(p)}$ is a phase-dependent base momentum. The direction gate suppresses extrapolation when consecutive anchor residuals diverge in orientation (negative cosine similarity is clamped to zero), while the magnitude gate attenuates the correction when residual norms change rapidly, preventing over-extrapolation during transient dynamics. When both residuals are well-aligned and similar in magnitude, $\beta_p^{(b)}$ approaches $\beta_0^{(p)}$, granting the full momentum correction; otherwise the gate gracefully decays toward zero, falling back to the first-order scaled estimate.

\paragraph{Anchor-Guided Unconditional Blending.}
The final unconditional draft (Section~\ref{sec:draft}) blends the extrapolation baseline with the anchor-transferred estimate:
\begin{equation}
\label{eq:blend}
\hat{\mathbf{u}}_t = (1 - \omega)\, \hat{\mathbf{u}}_t^{\mathrm{base}} + \omega\, \tilde{\mathbf{u}}_t,
\end{equation}
where $\hat{\mathbf{u}}_t^{\mathrm{base}}$ is the extrapolation output from Eq.~\eqref{eq:predictor}, $\tilde{\mathbf{u}}_t = \mathbf{u}_a + \alpha_p (\hat{\mathbf{c}}_t - \mathbf{c}_a)$ is the anchor-transferred unconditional estimate using a phase-dependent transfer coefficient $\alpha_p$, and $\omega$ is a blending weight. This correction leverages the conditional–unconditional coupling inherent in CFG to produce more faithful unconditional drafts, which in turn yields more accurate guided approximations during both skip execution and counterfactual error measurement.

\subsection{Full Ablation Grids for Scheduler-Related Axes}
\label{app:full-ablation}

We provide the full grids underlying Table~\ref{tab:abl_sched} of the main paper. 

\begin{table}[h]
\centering
\caption{Full grid for warm-up steps $W$.}
\label{tab:abl_sched_full_a}
\small
\begin{tabular}{c cc ccc}
\toprule
$W$ & Lat.(s)$\downarrow$ & Spd.$\uparrow$ & PSNR$\uparrow$ & SSIM$\uparrow$ & LPIPS$\downarrow$ \\
\midrule
5  & 42.73 & 2.29$\times$ & 23.68 & 0.829 & 0.123 \\
6  & 44.17 & 2.19$\times$ & 23.51 & 0.810 & 0.129 \\
7  & 43.12 & 2.26$\times$ & 24.40 & 0.850 & 0.110 \\
8  & 44.42 & 2.18$\times$ & 24.05 & 0.840 & 0.112 \\
9  & 43.72 & 2.22$\times$ & 24.51 & 0.862 & 0.104 \\
12 & 46.82 & 2.06$\times$ & 25.97 & 0.882 & 0.086 \\
14 & 49.03 & 1.97$\times$ & 26.34 & 0.887 & 0.078 \\
\bottomrule
\end{tabular}
\end{table}

\begin{table}[h]
\centering
\caption{Full grid for error base threshold $\theta_p$.}
\label{tab:abl_sched_full_b}
\small
\begin{tabular}{c cc ccc}
\toprule
$\theta_p$ & Lat.(s)$\downarrow$ & Spd.$\uparrow$ & PSNR$\uparrow$ & SSIM$\uparrow$ & LPIPS$\downarrow$ \\
\midrule
0.0325 & 53.77 & 1.81$\times$ & 27.20 & 0.904 & 0.070 \\
0.0350 & 54.06 & 1.79$\times$ & 26.58 & 0.892 & 0.076 \\
0.0375 & 50.29 & 1.94$\times$ & 26.04 & 0.888 & 0.082 \\
0.0400 & 50.42 & 1.91$\times$ & 25.48 & 0.870 & 0.090 \\
0.0425 & 47.21 & 2.06$\times$ & 25.58 & 0.880 & 0.088 \\
0.0475 & 45.06 & 2.16$\times$ & 24.93 & 0.868 & 0.098 \\
0.0600 & 41.75 & 2.31$\times$ & 24.05 & 0.843 & 0.114 \\
0.0700 & 39.35 & 2.45$\times$ & 23.65 & 0.832 & 0.123 \\
\bottomrule
\end{tabular}
\end{table}

\begin{table}[h]
\centering
\caption{Full grid for phase ratio $(\rho_e,\rho_l)$. The full 11-point grid confirms low sensitivity: PSNR varies by $<0.55$\,dB across all configurations.}
\label{tab:abl_sched_full_c}
\small
\begin{tabular}{c cc ccc}
\toprule
$(\rho_e,\rho_l)$ & Lat.(s)$\downarrow$ & Spd.$\uparrow$ & PSNR$\uparrow$ & SSIM$\uparrow$ & LPIPS$\downarrow$ \\
\midrule
(0.100, 0.86) & 45.17 & 2.14$\times$ & 24.70 & 0.860 & 0.100 \\
(0.125, 0.84) & 44.48 & 2.18$\times$ & 25.13 & 0.874 & 0.095 \\
(0.125, 0.88) & 44.29 & 2.19$\times$ & 25.11 & 0.873 & 0.095 \\
(0.150, 0.80) & 45.15 & 2.14$\times$ & 24.61 & 0.859 & 0.102 \\
(0.150, 0.82) & 44.46 & 2.18$\times$ & 25.00 & 0.872 & 0.097 \\
(0.150, 0.85) & 45.12 & 2.14$\times$ & 24.61 & 0.859 & 0.102 \\
(0.150, 0.88) & 44.28 & 2.19$\times$ & 24.99 & 0.871 & 0.097 \\
(0.150, 0.90) & 45.17 & 2.14$\times$ & 24.59 & 0.859 & 0.101 \\
(0.175, 0.84) & 44.47 & 2.18$\times$ & 25.06 & 0.873 & 0.096 \\
(0.175, 0.88) & 44.28 & 2.19$\times$ & 25.05 & 0.873 & 0.096 \\
(0.200, 0.86) & 45.14 & 2.14$\times$ & 24.65 & 0.859 & 0.100 \\
\bottomrule
\end{tabular}
\end{table}

\subsection{Case Studies}
\label{app:case_studies}

We provide additional qualitative comparisons covering all baselines (PAB, TeaCache, FasterCache, EasyCache, and AlignCache) on representative VBench prompts across two backbones, complementing the main-paper Figure~\ref{fig:case_study}.

\begin{figure}[htbp]
    \centering
    \includegraphics[width=\textwidth]{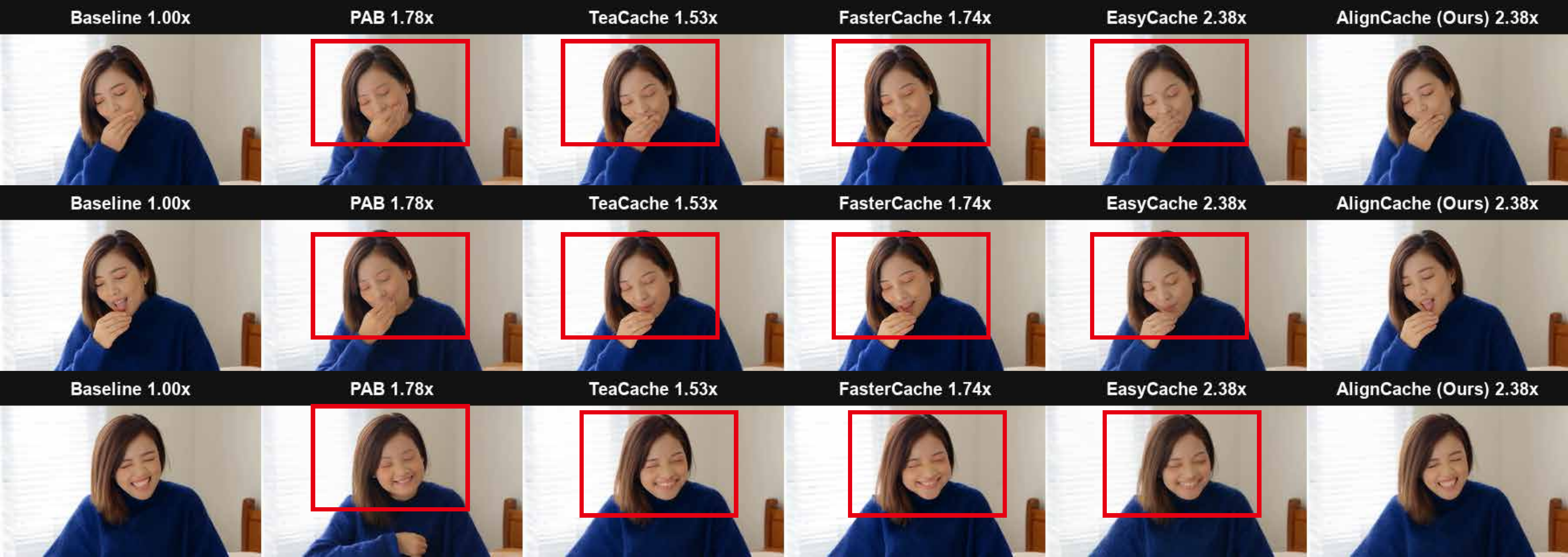}
    \caption{\textbf{Qualitative comparison} on \emph{``A person is kissing.''} generated by Wan2.1-1.3B. Red boxes highlight facial regions where PAB, TeaCache, FasterCache, and EasyCache introduce distortions in lip contour, skin tone, or identity-bearing details, while AlignCache preserves facial geometry and expression cues at the same $2.38\times$ speedup as EasyCache.}
\label{fig:case_kissing}
\end{figure}

Figure~\ref{fig:case_kissing} compares all methods on \emph{``A person is kissing.''} generated by Wan2.1-1.3B. Facial regions and lip contours are highly sensitive to caching-induced errors. PAB and TeaCache produce visible facial distortions, FasterCache exhibits mild ghosting, and EasyCache smooths out subtle expression cues at $2.38\times$. AlignCache, at the same $2.38\times$ speedup, faithfully preserves facial geometry, lip contour, and skin texture (red boxes).

\begin{figure}[htbp]
    \centering
    \includegraphics[width=\textwidth]{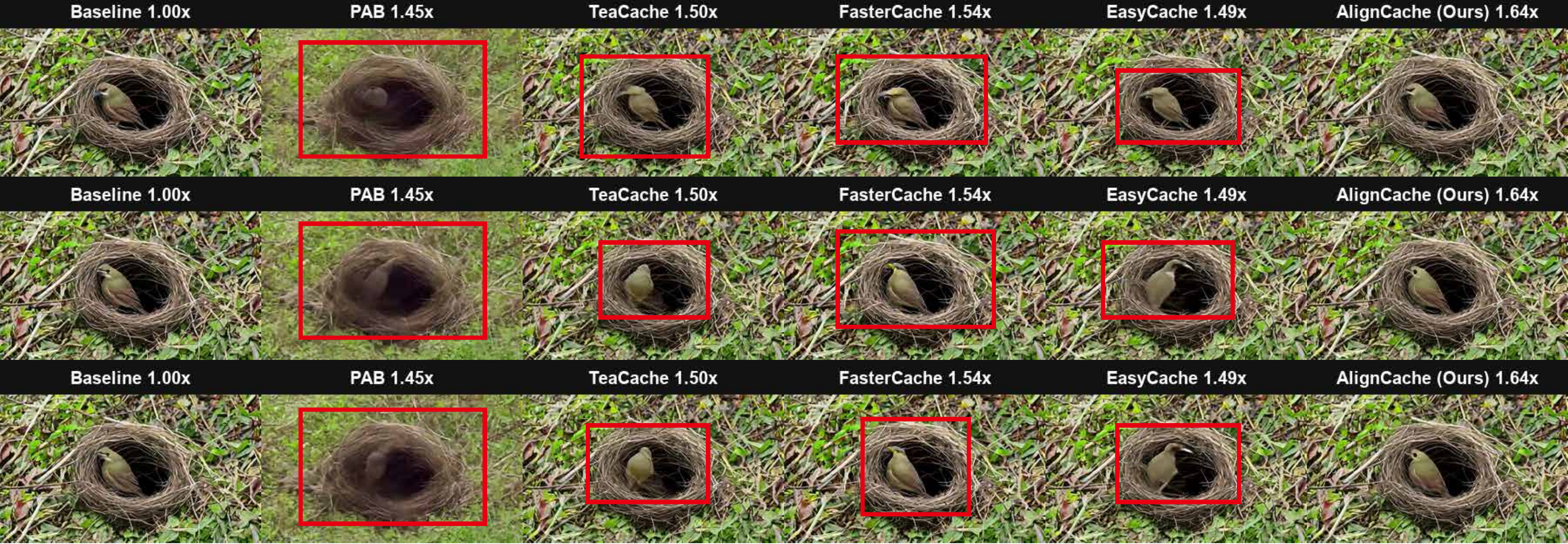}
    \caption{\textbf{Qualitative comparison} on \emph{``A bird building a nest from twigs and leaves.''} generated by Open-Sora-1.2. Red boxes mark high-frequency texture regions (nest twigs, foliage, feathers) where competing methods produce blurring, structural collapse, or color drift, whereas AlignCache retains fine woven structure and feather details at the highest $1.64\times$ speedup among compared methods.}
\label{fig:case_bird}
\end{figure}

Figure~\ref{fig:case_bird} reports a comparison on \emph{``A bird building a nest from twigs and leaves.''} generated by Open-Sora-1.2, a prompt dominated by high-frequency natural textures. Competing methods produce blurred twig boundaries, texture flattening, or color drift on the nest rim, while AlignCache, at the highest $1.64\times$ speedup, recovers the woven twig structure and fine feather details (red boxes). This is consistent with our analysis in Section~\ref{sec:case_studies}: high-frequency textures are where direct residual reuse incurs the largest error, and $E_t^{\mathrm{obs}}$ enables AlignCache to detect and correct such regions in real time.


\subsection{Broader Impacts}
\label{app:impact}

AlignCache reduces the inference cost of video diffusion models, lowering the energy footprint of large-scale deployment and improving accessibility for researchers with limited GPU resources. As a training-free acceleration layer, it inherits the dual-use risks of the underlying generators, faster inference could lower the barrier for misuse such as deepfakes or disinformation, and any biases of the base model are preserved. We therefore recommend deploying AlignCache alongside existing upstream safeguards, including content moderation, provenance watermarking, and the release policies of the original video generators.




\newpage
\input{checklist.tex}

\end{document}

%% file: checklist.tex
\section*{NeurIPS Paper Checklist}

\begin{enumerate}

\item {\bf Claims}
    \item[] Question: Do the main claims made in the abstract and introduction accurately reflect the paper's contributions and scope?
    \item[] Answer: \answerYes{} 
    \item[] Justification: The abstract and introduction clearly state the problem setting, the proposed AlignCache framework, and the claimed speed–quality trade-offs. These claims are supported by experiments on three video diffusion models and are scoped to training-free cache-based acceleration for video diffusion models.

\item {\bf Limitations}
    \item[] Question: Does the paper discuss the limitations of the work performed by the authors?
    \item[] Answer: \answerYes{} 
    \item[] Justification: Section 6 discusses remaining open directions and limitations, including prompt-independent calibration, warm-up overhead, and extension to long-form video generation and multimodal diffusion pipelines.

\item {\bf Theory assumptions and proofs}
    \item[] Question: For each theoretical result, does the paper provide the full set of assumptions and a complete (and correct) proof?
    \item[] Answer: \answerNA{} 
    \item[] Justification: The paper does not present formal theoretical results, theorems, or proofs. The equations are used to define the proposed scheduling, calibration, and draft synthesis procedures.

    \item {\bf Experimental result reproducibility}
    \item[] Question: Does the paper fully disclose all the information needed to reproduce the main experimental results of the paper to the extent that it affects the main claims and/or conclusions of the paper (regardless of whether the code and data are provided or not)?
    \item[] Answer: \answerYes{} 
    \item[] Justification: The paper describes the algorithmic components in Section 4 and reports the evaluated models, metrics, sampling settings, hyperparameters, prompt/sample counts, seeds, and GPU platforms in Section 5.1 and Appendix A.1. The paper also provides an anonymized code URL including scripts and instructions.

\item {\bf Open access to data and code}
    \item[] Question: Does the paper provide open access to the data and code, with sufficient instructions to faithfully reproduce the main experimental results, as described in supplemental material?
    \item[] Answer: \answerYes{} 
    \item[] Justification: The paper provides an anonymized code URL and uses public video diffusion backbones, public baselines where available. The released repository provides the implementation and reproduction instructions for the reported results.

\item {\bf Experimental setting/details}
    \item[] Question: Does the paper specify all the training and test details (e.g., data splits, hyperparameters, how they were chosen, type of optimizer) necessary to understand the results?
    \item[] Answer: \answerYes{} 
    \item[] Justification: Section 5.1 and Appendix A.1 specify the baselines, video generation models, resolutions, denoising steps, metrics, prompt/sample counts, seeds, hyperparameters, phase boundaries, and GPU platforms. Since the method is training-free, optimizer and training-split details are not applicable.

\item {\bf Experiment statistical significance}
    \item[] Question: Does the paper report error bars suitably and correctly defined or other appropriate information about the statistical significance of the experiments?
    \item[] Answer: \answerNA{} 
    \item[] Justification: Following the standard evaluation protocol of training-free diffusion acceleration works (e.g., TeaCache, FasterCache, EasyCache), we report aggregated metrics over multiple prompts and seeds rather than per-run error bars. Since our method is deterministic given a fixed seed and the relative ordering of methods is consistent across all evaluated prompts and backbones, formal significance testing is not the primary mode of comparison in this line of work.

\item {\bf Experiments compute resources}
    \item[] Question: For each experiment, does the paper provide sufficient information on the computer resources (type of compute workers, memory, time of execution) needed to reproduce the experiments?
    \item[] Answer: \answerYes{} 
    \item[] Justification: The paper reports GPU types and per-method latency/speedup, including RTX PRO 6000-96GB, H100-80GB, and Blackwell/Hopper comparisons. 
    
\item {\bf Code of ethics}
    \item[] Question: Does the research conducted in the paper conform, in every respect, with the NeurIPS Code of Ethics \url{https://neurips.cc/public/EthicsGuidelines}?
    \item[] Answer: \answerYes{} 
    \item[] Justification: The research proposes a training-free inference acceleration method evaluated on existing video generation models and benchmarks. We are not aware of any aspect of the work that deviates from the NeurIPS Code of Ethics.

\item {\bf Broader impacts}
    \item[] Question: Does the paper discuss both potential positive societal impacts and negative societal impacts of the work performed?
    \item[] Answer: \answerYes{} 
    \item[] Justification: A broader-impact discussion is provided at the Appendix~\ref{app:impact}, covering efficiency/accessibility benefits, dual-use risks inherited from base generators, and mitigation via upstream safeguards.
    
\item {\bf Safeguards}
    \item[] Question: Does the paper describe safeguards that have been put in place for responsible release of data or models that have a high risk for misuse (e.g., pre-trained language models, image generators, or scraped datasets)?
    \item[] Answer: \answerNA{} 
    \item[] Justification: The paper releases an acceleration method for existing video diffusion systems, while safeguards for the underlying generators are outside the scope of this work.

\item {\bf Licenses for existing assets}
    \item[] Question: Are the creators or original owners of assets (e.g., code, data, models), used in the paper, properly credited and are the license and terms of use explicitly mentioned and properly respected?
    \item[] Answer: \answerYes{} 
    \item[] Justification: The paper cites the existing video generators' versions, baselines, and evaluation benchmark used in the experiments.

\item {\bf New assets}
    \item[] Question: Are new assets introduced in the paper well documented and is the documentation provided alongside the assets?
    \item[] Answer: \answerYes{} 
    \item[] Justification: The paper introduces and releases the AlignCache code as a new asset through an anonymized repository. Documentation and reproduction instructions are provided with the repository, and the method itself is described in Sections 4 and A.1.

\item {\bf Crowdsourcing and research with human subjects}
    \item[] Question: For crowdsourcing experiments and research with human subjects, does the paper include the full text of instructions given to participants and screenshots, if applicable, as well as details about compensation (if any)? 
    \item[] Answer: \answerNA{} 
    \item[] Justification: The paper does not involve crowdsourcing, user studies, or research with human subjects.

\item {\bf Institutional review board (IRB) approvals or equivalent for research with human subjects}
    \item[] Question: Does the paper describe potential risks incurred by study participants, whether such risks were disclosed to the subjects, and whether Institutional Review Board (IRB) approvals (or an equivalent approval/review based on the requirements of your country or institution) were obtained?
    \item[] Answer: \answerNA{} 
    \item[] Justification: The paper does not involve human subjects, so IRB or equivalent approval is not applicable.

\item {\bf Declaration of LLM usage}
    \item[] Question: Does the paper describe the usage of LLMs if it is an important, original, or non-standard component of the core methods in this research? Note that if the LLM is used only for writing, editing, or formatting purposes and does \emph{not} impact the core methodology, scientific rigor, or originality of the research, declaration is not required.
    \item[] Answer: \answerNA{} 
    \item[] Justification: The core method development in this research does not involve LLMs as an important, original, or non-standard component. Any possible use of LLMs for writing, editing, or formatting would not affect the scientific methodology, rigor, or originality of the work.

\end{enumerate}